\documentclass{AVEC_FP}
\usepackage{gensymb}
\usepackage{graphicx}
\usepackage{subcaption}
\usepackage{booktabs}
\usepackage{multirow}
\usepackage{siunitx}
\usepackage{dashrule}
\usepackage{bm}
\usepackage{lipsum}
\usepackage{doi}

\newcommand{\Figure}{Fig.~}
\newcommand{\Table}{Table~}
\newcommand{\Section}{Section~}
\newcommand{\Appendix}{Appendix~}
\newcommand{\CG}{CG}
\newcommand{\vxin}{\dot{v}_{x}}
\begin{document}
\twocolumn[
\begin{titlepage}
    \papertitle{Short-Horizon Position Accuracy of Single-Track Models: Implications for Motion Planning of Autonomous Vehicles}
    \authors{Aron J. Aertssen\textsuperscript{1)}, Lars A.T.H. van Alen\textsuperscript{1)}, Igo J.M. Besselink\textsuperscript{1)}, Rudolf G.M. Huisman\textsuperscript{1,2)}, René M.J.G. van de Molengraft\textsuperscript{1)}}
    \contactinfo{1) Department of Mechanical Engineering, Eindhoven University of Technology, Groene Loper 3, 5612 AE Eindhoven, The Netherlands, 2) Safety \& Driver Controls Group, Vehicle Development, DAF Trucks N.V.,  Hugo van der Goeslaan 1, 5643 TW Eindhoven, The Netherlands \\
    (Corresponding author: \href{mailto:e-mail@address}{a.j.aertssen@tue.nl})}
    \receivedaccepted{M, D, YYYY}{M, D, YYYY}
    \myabstract{
    Accurate and computationally efficient vehicle models are essential for motion planning of autonomous vehicles, where positional accuracy directly affects trajectory feasibility and safety. However, the positional accuracy has not been systematically evaluated against real measurements. Therefore, this paper compares the short-horizon positional accuracy of three single-track vehicle models against vehicle measurements across various driving maneuvers. Model parameters are identified through dedicated experiments with the instrumented test vehicle. Rather than identifying a single best model, this work aims to provide insight into the trade-offs between model complexity, parameterization quality, and positional accuracy for informed model selection in Model Predictive Control applications.}
\end{titlepage}
]

\thispagestyle{fancy}

\section{Introduction}
Autonomous vehicles are increasingly being deployed across a wide range of applications, from driverless taxis to autonomous buses and trucks. As part of their driving task, these vehicles must plan and execute safe, comfortable, and efficient trajectories in complex, dynamic environments, avoiding collisions with other road users and infrastructure while respecting road and lane boundaries.

Planning and control algorithms typically rely on vehicle models to generate trajectories and compute the control actions (e.g., steering, acceleration and braking) required to execute them. A widely used framework that combines trajectory planning and execution is Model Predictive Control (MPC)\cite{10.1016/j.robot.2024.104630, 10.1109/TIV.2016.2578706, 10.1016/j.arcontrol.2022.11.001}. In MPC, the control action is obtained by solving, at each sampling instant, an open-loop finite-horizon optimal control problem initialized at the current plant state. The optimisation yields a finite sequence of control inputs, of which only the first is applied to the plant\cite{2020_Rawlings_Modelpredictivecontroltheorycomputationdesign}. The plant model, i.e., the vehicle model, is embedded in this framework and defines the predicted trajectory over the horizon and shapes the requested control actions. Consequently, the fidelity of the vehicle model is key for closed-loop performance, as model mismatch degrades both trajectory tracking accuracy and control quality.

\begin{figure}[t]
    \centering
    \begin{subfigure}[t]{0.99\linewidth}
        \centering
        \includegraphics[width=0.7\linewidth]{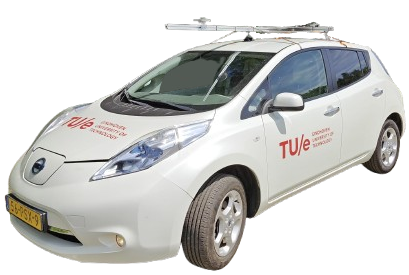}
        \caption{}
        \label{fig:Leaf}
    \end{subfigure}\\
    \begin{subfigure}[t]{0.49\linewidth}
        \centering
        \includegraphics[width=1\linewidth]{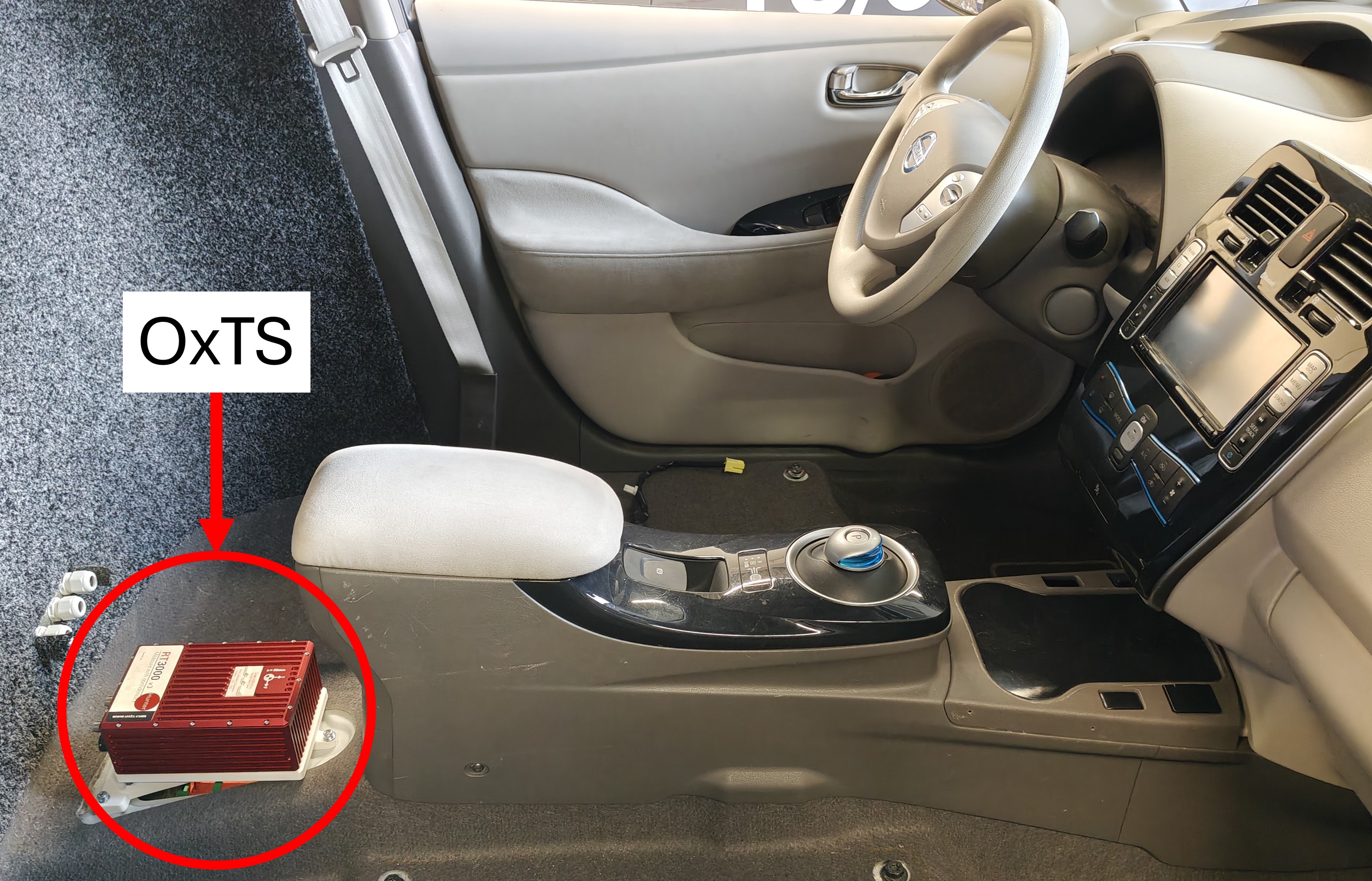}
        \caption{}
        \label{fig:OxTS_position}
    \end{subfigure}
    \hfill
    \begin{subfigure}[t]{0.49\linewidth}
        \centering
        \includegraphics[width=1\linewidth]{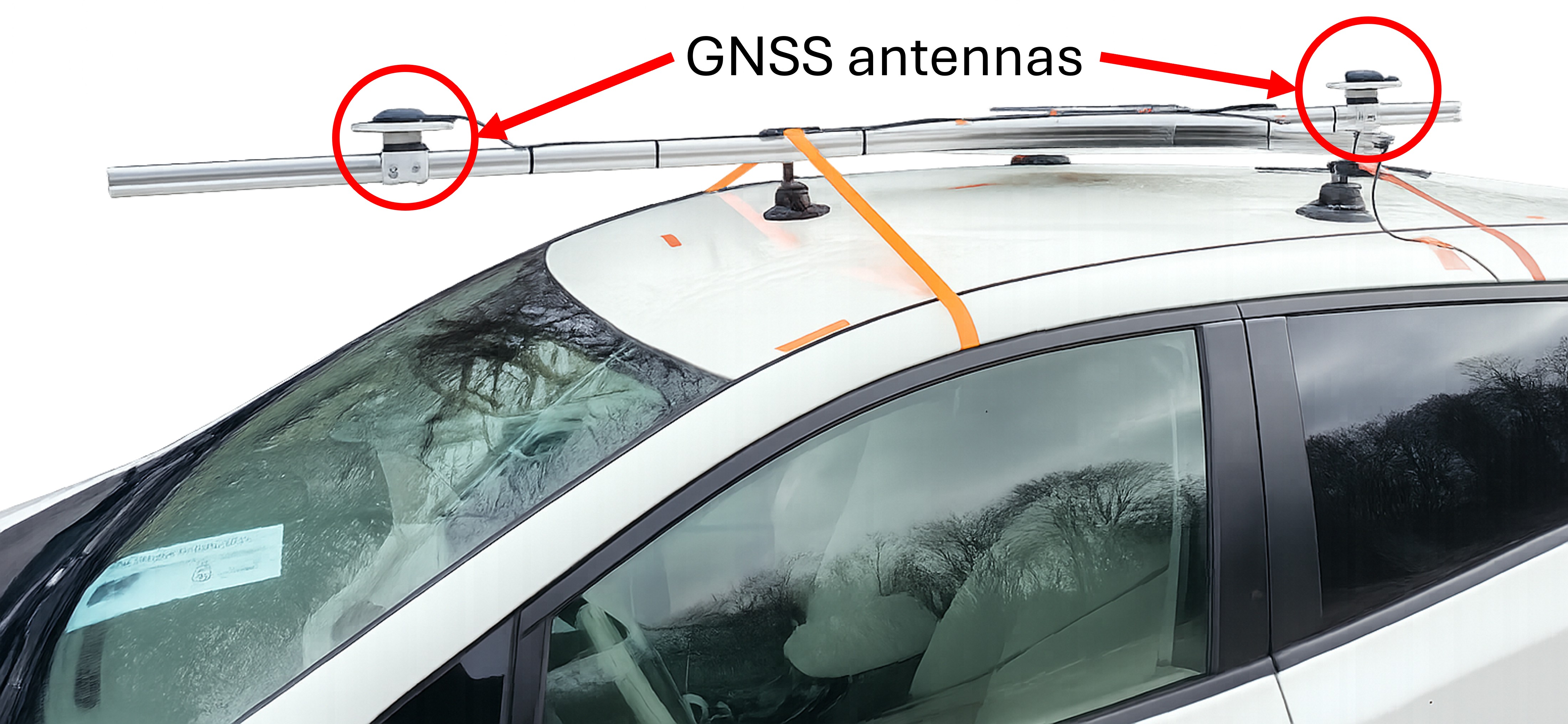}
        \caption{}
        \label{fig:antenna_frame}
    \end{subfigure}
    \caption{(a) Nissan Leaf ZE0 (2011), instrumented with an Oxford Technical Solutions (OxTS) GNSS-INS and CSS Electronics CANedge2 data logger, used to record vehicle measurements, (b) mounting location of the OxTS, and (c) frame mounted onto the roof for the OxTS GNSS antennas.}
    \label{fig:vehicle}
\end{figure}

At the same time, these algorithms are subject to computational constraints, since they must be able to respond quickly to disturbances, for example when another vehicle enters the ego lane. Consequently, high-fidelity vehicle models are often too computationally expensive for online use. Therefore, many methods in the literature rely on the single-track (bicycle) vehicle model, such as \cite{10.1016/j.ifacol.2019.08.085, 10.1109/IVS.2015.7225830, 10.1109/TIV.2019.2938102}. In this model, the left and right tires of each axle are lumped into a single equivalent tire located on the vehicle centerline. There exist variations of the single-track model, for example whether tire slip is neglected (kinematic), modeled using a linear tire model, or represented by a nonlinear tire model such as Pacejka's Magic Formula\cite{2012_Pacejka_Tirevehicledynamics}. As a result, these variations of the single-track model offer a trade-off between accuracy and computational cost.

Historically, position accuracy has received less attention in vehicle dynamics research since vehicle models are often evaluated based on acceleration and yaw-rate responses\cite{10.1080/00423114.2013.868500}. However, with the increasing deployment of autonomous vehicles in everyday traffic, the position accuracy of vehicle models used for planning and control becomes essential for safe, comfortable, and efficient driving. Position accuracy is especially important at low speeds, where tight tolerances are required for tasks such as parking maneuvers of cars or reverse docking of trucks.

To this end, Kong et al.\cite{10.1109/IVS.2015.7225830} have compared kinematic and linear-tire dynamic single-track models using experimental vehicle measurements and report similar open-loop errors over short horizons. Polack et al.\cite{10.1109/IVS.2017.7995816} compare a kinematic bicycle model with a 9-degree-of-freedom vehicle model in simulation and show that the kinematic model becomes increasingly inaccurate at higher speeds and steering angles because it does not capture slip-induced understeer. Ren et al.\cite{10.4271/2014-01-0841} compare linear and nonlinear bicycle models against a high fidelity simulation. They show that the linear bicycle model matches the reference trajectory well at low speed, while the nonlinear bicycle model remains accurate over a wider operating range and during more severe maneuvers. Overall, their results indicate that nonlinear tire behavior and load-transfer effects become important once the vehicle operates beyond the linear regime. 

The goal of this paper is to provide insight into single-track vehicle model choices suitable for MPC applications. To this end, we evaluate the positional accuracy of three single-track model variants against real passenger car measurements over a fixed, short prediction horizon across a range of driving maneuvers, representative of the receding-horizon predictions made in MPC. The variants considered are a kinematic model, a dynamic model with linear tires, and a dynamic model with nonlinear Magic Formula tires and quasi-static longitudinal load transfer. In doing so, we aim to bridge the gap between the autonomous driving and vehicle dynamics communities by providing insight into model trade-offs in MPC applications. Beyond the evaluated variants, we discuss additional modeling considerations so an appropriate model for their intended operating conditions can be selected.

The remainder of this paper is structured as follows. First, the vehicle measurement platform is discussed in \Section\ref{sec:vehicle_measurement_platform}. The considered vehicle models are described in \Section\ref{sec:veh_mod}, followed by the comparison of the models to the measurements in \Section\ref{sec:results}. In \Section\ref{sec:discussion}, practical considerations for selecting an appropriate vehicle model for motion planning are discussed. Finally, conclusions and future work are discussed in \Section\ref{sec:conc}.

\begin{table}[t]
    \centering
    \caption{Vehicle parameters for the Nissan Leaf ZE0 in \Figure\ref{fig:Leaf} together with simulation parameters.}
    \label{tab:veh_par}
    \begin{tabular}{cccc}
        \textbf{Definition} & \textbf{Notation} & \textbf{Value} & \textbf{Unit} \\ \hline
        front axle~-~\CG~distance & $l_0$ & 1.176 & m \\
        rear axle~-~\CG~distance & $l_1$ & 1.524 & m \\
        \CG~-~OxTS distance & $l_{OxTS}$ & 0.398 & m \\ 
        wheelbase & $l$ & 2.7 & m  \\
        vehicle mass (incl. driver) & $m$ & 1590 & kg \\
        height of \CG & $h_{\CG}$ & 0.443 & m \\
        steer ratio & $i_s$ & 17.3 & - \\
        yaw inertia & $I_{zz}$ & 2925 & kg m$^2$ \\
        front cornering stiffness & $C_{\alpha,0}$ & 70000 & N/rad \\
        rear cornering stiffness & $C_{\alpha,1}$ & 106000 & N/rad \\
        threshold for low speeds& $v_{\epsilon}$ & 3 & m/s
    \end{tabular}
\end{table}

\section{Vehicle Measurement Platform} \label{sec:vehicle_measurement_platform}
\subsection{Instrumentation}

Measurements are conducted using a 2011 Nissan Leaf ZE0 (\Figure\ref{fig:Leaf}). An Oxford Technical Solutions (OxTS) RT3000 v3 is mounted behind the front seats on the vehicle centerline (\Figure\ref{fig:OxTS_position}), with two GNSS antennas installed on the roof (\Figure\ref{fig:antenna_frame}). The OxTS RT3000 v3 is a combined GNSS-INS system that fuses real-time kinematic (RTK) GPS with inertial measurements to provide high-accuracy position, velocity, and acceleration data. Using RTK corrections via NTRIP, it achieves 0.01~m position accuracy\cite{2025__RT3000v3OXTS}. Vehicle signals, including steering wheel angle, are acquired via the OBD-II port of the instrumented vehicle using a CSS Electronics CANedge2 data logger\cite{__CANedge22xCANBusDataLoggerSDWiFi}. To enable synchronization between the separate OxTS and CANedge2 datasets, selected OxTS signals, such as the longitudinal velocity and yaw angle, are transmitted via CAN to the CANedge2. During post-processing, these shared signals are overlaid to determine the time offset and subsequently align the datasets. 

\subsection{Parameter Identification} \label{sec:vehicle_measurement_platform_params}
The vehicle parameters required for the models are listed in \Table\ref{tab:veh_par}. The vehicle mass is determined using a four-scale setup, measuring the mass per wheel and summing to obtain the total mass $m$. The longitudinal \CG~position ($l_0$, $l_1$) follows from the front-to-rear axle mass distribution. By lifting the front axle to several pitch angles, measuring the resulting rear axle load change, and applying the ISO~10392 method described in \cite{10.4271/2026-26-0513}, the \CG~height $h_{\CG}$ is obtained by averaging the value over all pitch angles. The steer ratio $i_s$ is identified by placing the front wheels on steering turntables, steering the steering wheel in approximately 45{\degree} increments (actual value is logged on the CANedge2), and fitting a linear relation between steering wheel angle $\delta_h$ and the averaged left/right wheel steer angle. A constant steering wheel angle offset of $-2.7\degree$ was identified to align the simulated yaw rate response with a set of measurements for straight-line driving scenarios as well as possible. The yaw inertia $I_{zz}$ is taken from\cite{10.1016/j.ifacol.2021.08.575}, which uses a comparable Nissan Leaf ZE0 for steering dynamics identification. 

The cornering stiffnesses $C_{\alpha i}$, where $i=0$ corresponds to the front axle and $i=1$ to the rear axle, are identified from experimental driving data using a least-squares fit of the linear tire model. Low-speed and near-straight samples are discarded to ensure sufficient excitation, and high lateral-acceleration data are excluded ($|a_y| > 4~\mathrm{m/s^2}$) to restrict the fit to the linear tire regime. The front and rear slip angles $\alpha_i$ are computed from the velocity components at each axle (see \eqref{eq:sideslipangles}, \eqref{eq:wheelvelocities}, and \Figure\ref{fig:veh_model_wheel_velocities}), and the corresponding axle lateral forces $F_{yi}$ are obtained from the force and yaw moment balance of the dynamic single-track model. The cornering stiffnesses $C_{\alpha i}$ are then obtained as the slopes of a linear regression of $F_{yi}$ on $\alpha_{i}$ for $|\alpha_{i}| \leq 2^\circ$. With this approach, secondary compliance effects such as steering compliance and roll steer are implicitly absorbed into the identified cornering stiffnesses.

\begin{figure}[t]
    \centering
    \begin{subfigure}[t]{0.48\linewidth}
        \centering
        \includegraphics[width=\linewidth]{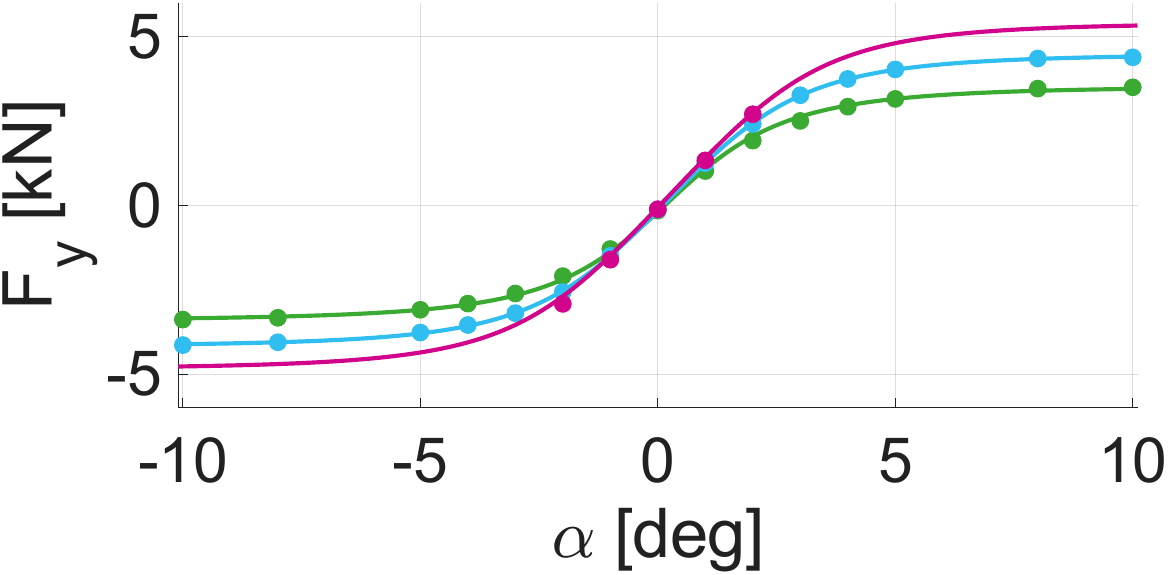}
        \caption{}
        \label{fig:measured_and_fitted_tire_forces_f}
    \end{subfigure}
    \hfill
    \begin{subfigure}[t]{0.48\linewidth}
        \centering
        \includegraphics[width=\linewidth]{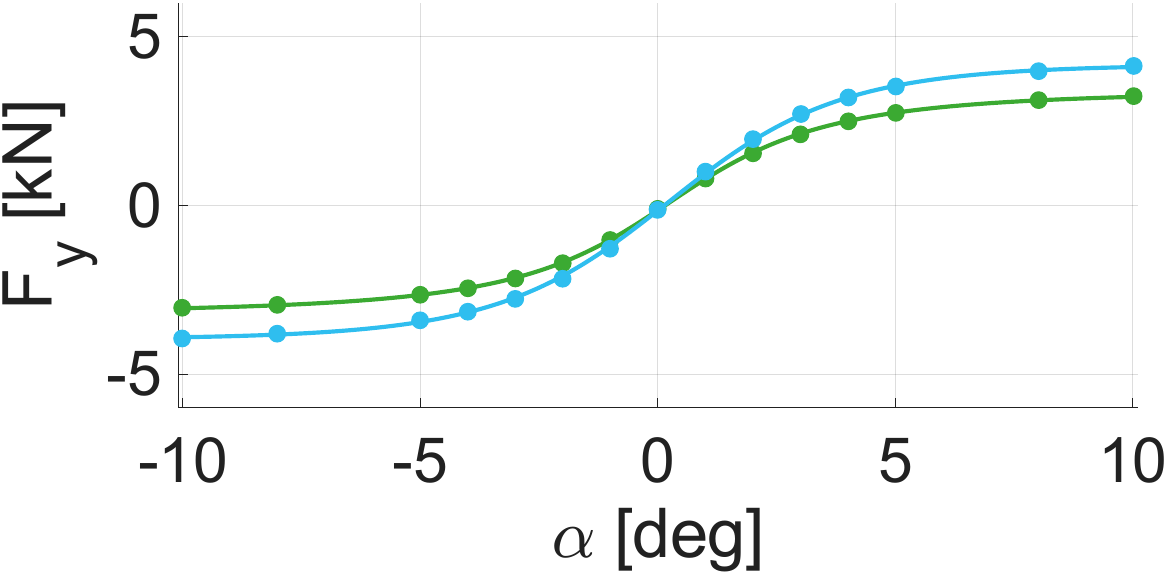}
        \caption{}
        \label{fig:measured_and_fitted_tire_forces_r}
    \end{subfigure}
    \caption{Measured steady-state lateral forces (\textbullet) and Magic Formula forces (\textcolor[rgb]{0,0,0}{\rule[0.5ex]{11pt}{1.5pt}}) for the front (a) and rear (b) tire for $F_z = 3300$\,N (\textcolor[rgb]{0.231,0.666,0.196}{\rule[0.5ex]{11pt}{1.5pt}}), $F_z = 4250$\,N (\textcolor[rgb]{0.184,0.745,0.937}{\rule[0.5ex]{11pt}{1.5pt}}), and $F_z = 5200$\,N (\textcolor[rgb]{0.819,0.015,0.545}{\rule[0.5ex]{11pt}{1.5pt}}).}
    \label{fig:measured_and_fitted_tire_forces}
\end{figure}

{The Magic Formula (MF) parameters of the front and rear tires are identified using the flat plank tire test facility of Eindhoven University of Technology\cite{2023_Aertssen_FlatPlankTireTesterEnhancementsTransientMeasurementsModelParametrization}. On this machine, a constant slip angle is applied at a constant axle height, and the steady-state lateral force is recorded. The lateral force characteristics for the two tires at various vertical loads are shown in \Figure\ref{fig:measured_and_fitted_tire_forces}, together with the fitted MF forces. The MF parameters are first fitted for the nominal vertical load, after which the vertical load dependency parameters are determined from the non-nominal results. The identified parameters are listed in \Table\ref{tab:MF_pars} in \Appendix\ref{sec:appendix_MF}.}

Since the MF equations purely describe the tire behavior, they do not include the aforementioned compliance effects. To reconcile both tire models, the cornering stiffness scaling factor $\lambda_{Ky\alpha i}$ of the MF is computed as
\begin{equation}
    \lambda_{Ky\alpha i} = \frac{C_{\alpha i}}{2K_{y\alpha i}},
\end{equation}
where the MF tire cornering stiffness $K_{y\alpha i}$ is evaluated at the nominal tire vertical force with $\lambda_{Ky\alpha i}=1$ using \eqref{eq:MF_cornering_stiffness} in \Appendix\ref{sec:appendix_MF}. The factor two is a result of $C_{\alpha i}$ referring to an axle and $K_{y \alpha i}$ to a single tire.

\subsection{GNSS-INS Sensor Offset Correction}
\begin{figure}[t]
    \centering
    \subcaptionbox{\label{fig:veh_model_kinematic}}{\includegraphics[width=0.48\columnwidth]{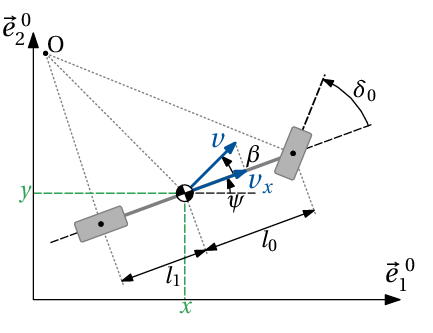}}%
    \hfill
    \subcaptionbox{\label{fig:veh_model_dynamic}}{\includegraphics[width=0.48\columnwidth]{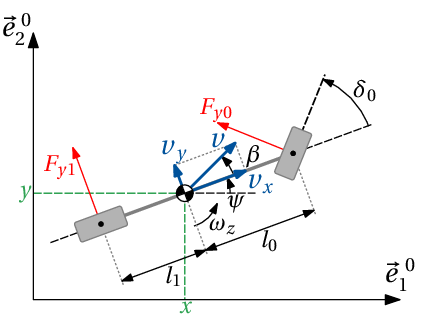}}
    \caption{Schematic representation of the (a) kinematic and (b) dynamic single-track model.}
    \label{fig:veh_models}
\end{figure}

Since the OxTS is mounted behind the vehicle's \CG~at the vehicle plane of symmetry, the measured velocities and accelerations need to be determined for the \CG:
\begin{subequations}
    \begin{align}
        v_{x,\CG} &= v_{x,\text{OxTS}} - \omega_z d_y, \\
        v_{y,\CG} &= v_{y,\text{OxTS}} + \omega_z d_x,\\
        a_{x,\CG} &= a_{x,\text{OxTS}} - \omega_z^2 d_x - \dot{\omega}_z d_y, \\
        a_{y,\CG} &= a_{y,\text{OxTS}} + \dot{\omega}_z d_x - \omega_z^2 d_y,
    \end{align}
\end{subequations}
where $d_y = 0$, and $d_x = l_{OxTS} > 0$ is the forward offset from the OxTS to the \CG~in the body frame (i.e., the sensor is mounted rearward of the \CG), the longitudinal and lateral velocity at the \CG~in the body frame are denoted by $v_x$ and $v_y$, respectively, $\omega_z$ is the yaw rate and $\dot{\omega}_z$ its time derivative. The position of the OxTS is given by
\begin{subequations} \label{eq:pos_OxTS}
    \begin{align}
        x_{OxTS} &= x - l_{OxTS} \cos{\left(\psi\right)},\\
        y_{OxTS} &= y - l_{OxTS} \sin{\left(\psi\right)},
    \end{align}
\end{subequations}
where $x$ and $y$ are the coordinates of the center of gravity (\CG) of the vehicle in the global Cartesian coordinate frame $\vec{e}^{\,0}$, as illustrated in \Figure\ref{fig:veh_models}, and $\psi$ is the yaw angle of the vehicle. The distance from the \CG~to the OxTS sensor is $l_{OxTS} > 0$, where the negative sign in~\eqref{eq:pos_OxTS} reflects that the sensor is mounted rearward of the \CG.

\section{Vehicle Models} \label{sec:veh_mod}
This work evaluates three single-track vehicle models: a kinematic single-track model (kinematic), a dynamic single-track model with linear tires (linear dynamic), and a dynamic single-track model with nonlinear Magic Formula tires and quasi-static longitudinal load transfer (nonlinear dynamic). The inputs for these three models are the steering angle $\delta_0$ at the front wheel and the time derivative of the OxTS forward velocity $\vxin$, leading to input vector $\bm{u}$ being equal to
\begin{equation}
    \bm{u} = \begin{bmatrix}
        \delta_0 & \vxin
    \end{bmatrix}^\top.
\end{equation}
The steering angle is determined from the measured steering wheel angle $\delta_h$ via a constant steering ratio $i_s$:
\begin{equation} \label{eq:steering}
    \delta_0= \frac{\delta_h}{i_s}.
\end{equation}
We assume that the rear axle is non-steered.

The input $\vxin$ is used rather than the OxTS longitudinal acceleration $a_{x,\text{OxTS}}$. The longitudinal acceleration measured by the OxTS includes a yaw-rate coupling term $\dot{\psi} v_y$, meaning it does not purely represent $\dot{v}_x$ as defined in the vehicle model. Additionally, measured acceleration values are commonly noisy and are influenced by pitch and roll effects. Therefore, $\vxin$ is obtained by applying a first-order Butterworth filter (5~Hz cutoff frequency) to the OxTS longitudinal velocity signal and subsequently numerically differentiating it.

The state vector $\bm{x}$ for each model is given by 
\begin{equation}
\bm{x} = 
\begin{bmatrix}
    x & y & v_x & v_y & \psi & \omega_z
\end{bmatrix}^\top,
\end{equation}
where $x$ and $y$ are the \CG~position coordinates, $\psi$ is the yaw angle, and $v_x$, $v_y$, and $\omega_z$ are the longitudinal velocity, lateral velocity, and yaw rate, all defined at the \CG.

\begin{figure}[t]
    \centering
    \includegraphics[width=0.48\columnwidth]{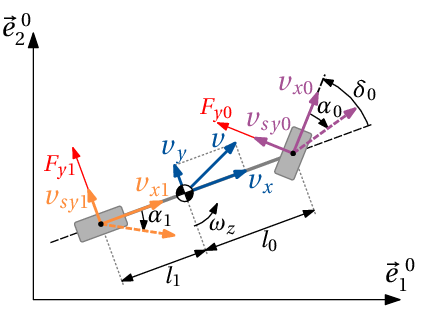}
    \caption{Velocities and side slip angles in the wheel frames.}
    \label{fig:veh_model_wheel_velocities}
\end{figure}

\subsection{Kinematic Model}
The kinematic model neglects tire slip and assumes rotation about an instantaneous center $O$ perpendicular to the rear-axle line, see \Figure\ref{fig:veh_model_kinematic}. The resulting curvature depends only on steering angle and wheelbase, so speed-dependent steering effects (e.g., understeer or oversteer) are not captured. The continuous-time equations of the kinematic vehicle model are given by
\begin{equation}\label{eq:kin_model}
\dot{\bm{x}}
=
\begin{bmatrix}
v_x \cos(\psi + \beta) \\
v_x \sin(\psi + \beta) \\
\vxin \\
0 \\
\dfrac{v_x}{l} \cos\left(\beta\right) \, \tan\left(\delta_0\right)\\
0 \\
\end{bmatrix},
\end{equation}
where $l = l_0 + l_1$ is the wheelbase and $\beta$ is the vehicle side slip angle, which in the kinematic model is fully determined by geometry:
\begin{equation}
    \beta = \arctan\!\left(\frac{l_1}{l}\tan\left(\delta_0\right)\right).
\end{equation}
Note that the kinematic model imposes no-slip constraints at both axles, so $v_y$ and $\omega_z$ are not independent states but can be algebraically determined using the forward velocity $v_x$, steering angle $\delta_0$ and their derivatives. The zero entries in \eqref{eq:kin_model} reflect that these rows carry no additional dynamic information and they are retained only to preserve a uniform state definition $\bm{x}$ across all three models.

\subsection{Dynamic Model}
The linear and nonlinear dynamic models include lateral and yaw dynamics about the center of gravity, and include tire slip angles, see \Figure\ref{fig:veh_model_dynamic}. The continuous-time equations of the dynamic model are given by
\begin{equation}
\dot{\bm{x}} = 
\begin{bmatrix}
v_x \cos\left(\psi\right) - v_y \sin\left(\psi\right)                                     \\
v_x \sin\left(\psi\right) + v_y \cos\left(\psi\right)                                     \\
\vxin                                                    \\
\frac{1}{m}\left(F_{y0} \cos\left(\delta_0 \right) + F_{y1} \right) - v_x \omega_z                    \\
\omega_z                                                        \\
\frac{1}{I_{zz}}\left( l_{0}\, F_{y0} \cos\left(\delta_0 \right) - l_{1}\, F_{y1} \right)           \\
\end{bmatrix}
\end{equation}
where $F_{y0}$ and $F_{y1}$ are the lateral tire forces of the front and rear axle, respectively. 

For the linear dynamic model, linear tire behavior with constant cornering stiffness $C_{\alpha i}$ is assumed, leading to
\begin{equation} \label{eq:lateral force}
    F_{yi} = C_{\alpha i} \alpha_i,
\end{equation}
where $\alpha_i$ is the tire side slip angle, with $i \in \{0,1\}$ denoting the front and rear axle, respectively. The side slip angle is given by
\begin{equation} \label{eq:sideslipangles}
    \alpha_i = \arctan\left(\frac{v_{syi}}{\max(|v_{xi}|, v_{\epsilon})}\right),
\end{equation}
where $v_{\epsilon} = 3$\,m/s is a threshold value to maintain numerical robustness at low speeds. To verify that $v_{\epsilon}$ does not significantly affect the results, a simulation was run with a timestep of 1~ms and compared against the 50~ms baseline, where the vehicle velocity is in the vicinity of $v_\epsilon$. The resulting trajectory errors are within approximately 5\% of the baseline values. 

To calculate the tire side slip angles, the velocities $v_{xi}$ and $v_{syi}$ defined in the wheel-fixed coordinate system (see \Figure\ref{fig:veh_model_wheel_velocities}) are determined by
\begin{subequations} \label{eq:wheelvelocities}
    \begin{align}
    v_{x0} &= v_x \cos\left(\delta_0\right) + \left(v_y + l_0\omega_z\right) \sin\left(\delta_0\right) \\
    v_{sy0} &= -v_x \sin\left(\delta_0\right) + \left(v_y + l_0\omega_z\right) \cos\left(\delta_0\right) \\
    v_{x1} &= v_x \\
    v_{sy1} &= v_y - l_1\omega_z.
\end{align}
\end{subequations}

\begin{figure}[t]
    \centering
    \includegraphics[width=0.48\linewidth]{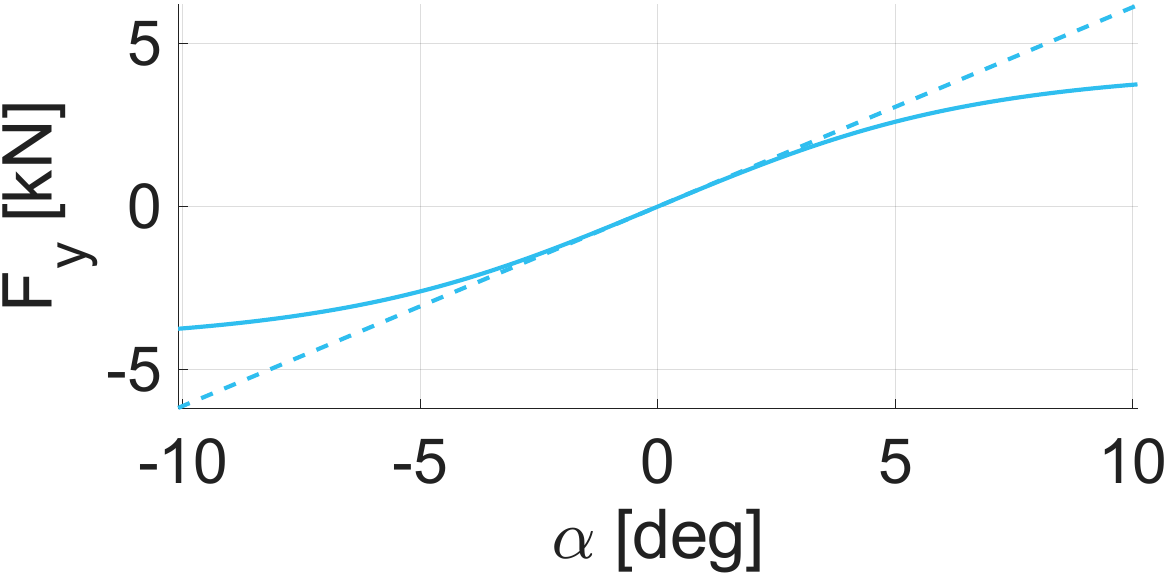}
    \caption{Lateral tire force as a function of the tire side slip angle for the linear (\textcolor[rgb]{0,0,0}{\hdashrule[0.5ex]{11pt}{1.5pt}{2pt 1pt}}) and nonlinear (\textcolor[rgb]{0,0,0}{\rule[0.5ex]{11pt}{1.5pt}}) tire model for the front tire at $F_z = 4250$\,N (\textcolor[rgb]{0.184,0.745,0.937}{\rule[0.5ex]{11pt}{1.5pt}}). Note that, compared to the forces in \Figure\ref{fig:measured_and_fitted_tire_forces}, the nonlinear MF force response is scaled using $\lambda_{K y \alpha i}$ as explained in \Section\ref{sec:vehicle_measurement_platform_params}.}
    \label{fig:comparison_linear_nonlinear_model}
\end{figure}

For the nonlinear dynamic model, lateral tire forces are computed using the nonlinear Magic Formula under the pure side-slip assumption. \Figure\ref{fig:comparison_linear_nonlinear_model} visualizes the linear and nonlinear force response, where the nonlinear forces decrease for increasing tire side slip angles. The lateral force at each axle depends on the vertical tire force, which varies with longitudinal acceleration due to the elevated center of gravity. Therefore, quasi-static longitudinal load transfer is included, making axle loads acceleration-dependent. The vertical force $F_{zi}$ for $i \in \{0,1\}$ is given by
\begin{subequations} \label{eq:load_transfer}
\begin{align}
    F_{z0} &= \frac{m}{l}\left( g l_{1} - a_x h_{\CG} \right),\\
    F_{z1} &= \frac{m}{l}\left( g l_{0} + a_x h_{\CG} \right),
\end{align}
\end{subequations}
where $h_{\CG}$ is the center-of-gravity height and $g$ is the gravitational acceleration. We assume that $a_x = \vxin$, thus neglecting the yaw motion components. As a consequence of the load transfer, acceleration and braking influence the side-slip response and can shift the vehicle balance from understeer toward neutral steer or oversteer. The lateral forces $F_{yi}$ are computed using the MF equations\cite{2012_Pacejka_Tirevehicledynamics} given in \eqref{eq:MF} in \Appendix\ref{sec:appendix_MF}, which incorporate the dependency on $F_{zi}$. Since the MF parameters in \Table\ref{tab:MF_pars} and the equations in \Appendix\ref{sec:appendix_MF} correspond to a single tire, whereas the single-track model represents each axle as one equivalent tire, the vertical load is halved prior to evaluating the MF, and the resulting lateral force is doubled to yield the axle-equivalent force. Note that lateral load transfer, which in reality causes the inner and outer tires to carry unequal vertical forces and thus generate different lateral forces, is not captured by this approach.

\begin{figure*}[t]
    \centering
    \begin{subfigure}[t]{0.23\linewidth}
        \centering
        \includegraphics[width=\linewidth]{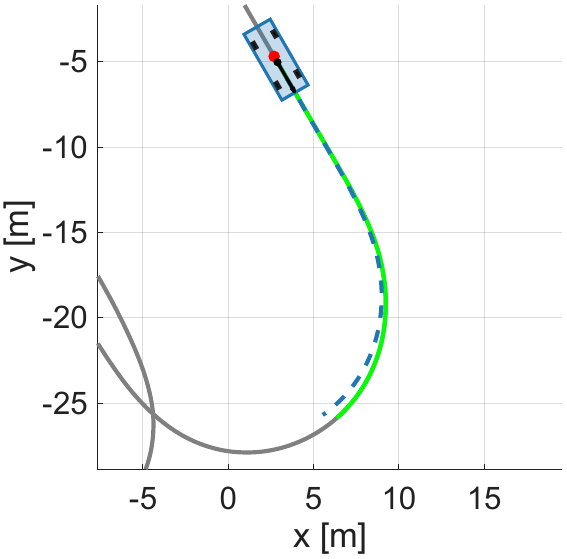}
        \label{fig:paperclip_traj_0}
    \end{subfigure}
    \hfill
    \begin{subfigure}[t]{0.23\linewidth}
        \centering
        \includegraphics[width=\linewidth]{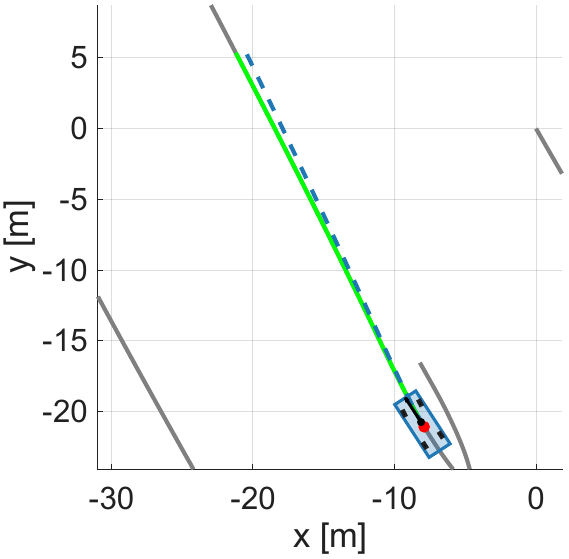}
        \label{fig:paperclip_traj_1}
    \end{subfigure}
    \hfill
    \begin{subfigure}[t]{0.23\linewidth}
        \centering
        \includegraphics[width=\linewidth]{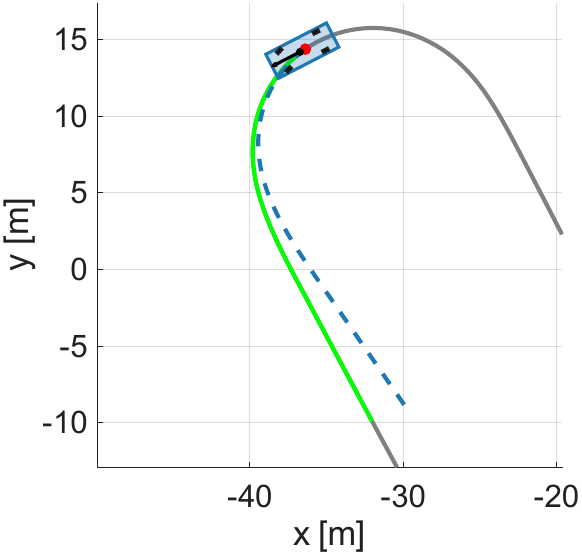}
        \label{fig:paperclip_traj_2}
    \end{subfigure}
    \hfill
    \begin{subfigure}[t]{0.23\linewidth}
        \centering
        \includegraphics[width=\linewidth]{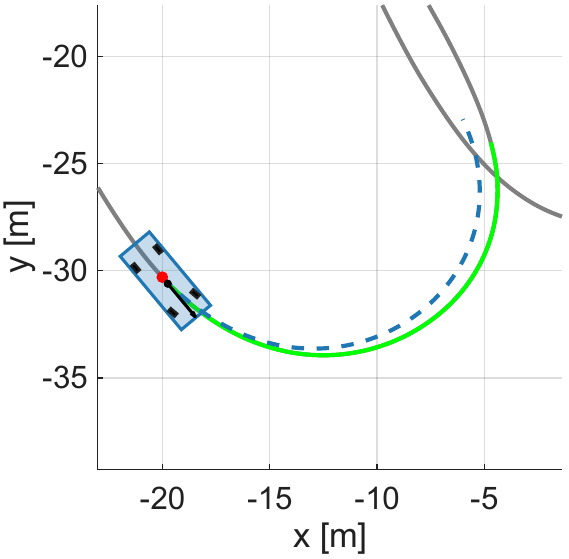}
        \label{fig:paperclip_traj_3}
    \end{subfigure}
    \vspace{0.5em}
    \begin{subfigure}[t]{0.23\linewidth}
        \centering
        \includegraphics[width=\linewidth]{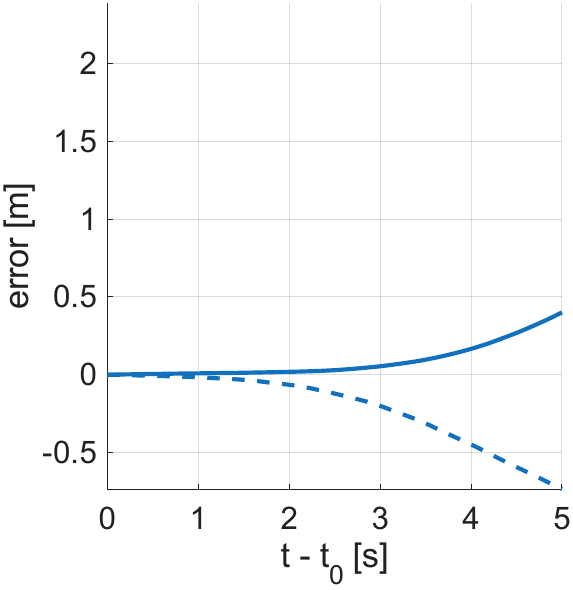}
        \caption{}
        \label{fig:paperclip_err_0}
    \end{subfigure}
    \hfill
    \begin{subfigure}[t]{0.23\linewidth}
        \centering
        \includegraphics[width=\linewidth]{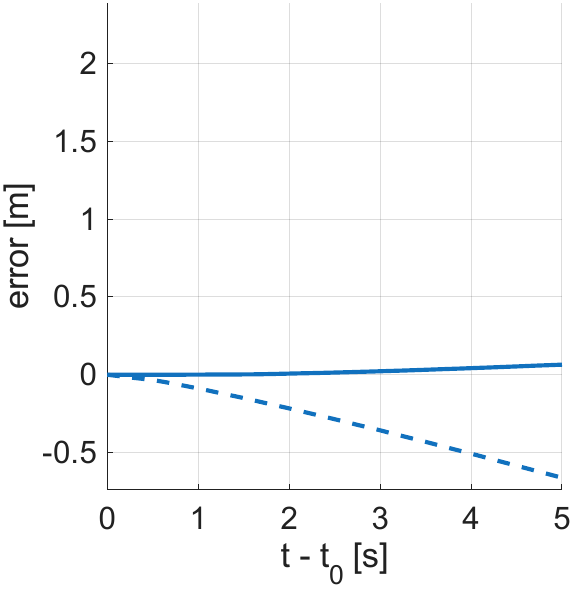}
        \caption{}
        \label{fig:paperclip_err_1}
    \end{subfigure}
    \hfill
    \begin{subfigure}[t]{0.23\linewidth}
        \centering
        \includegraphics[width=\linewidth]{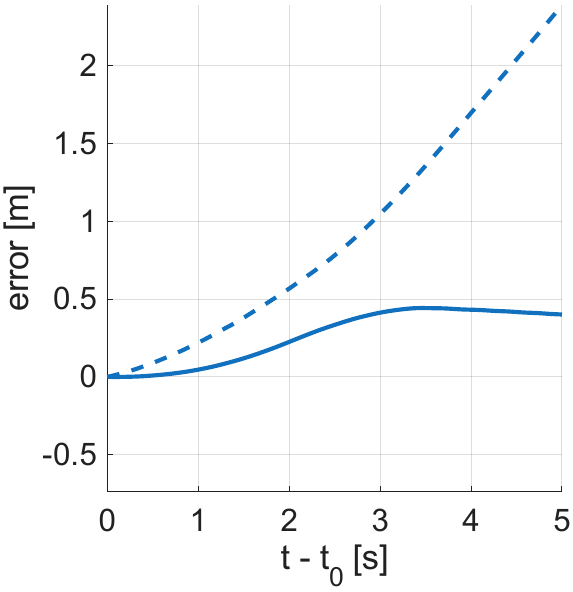}
        \caption{}
        \label{fig:paperclip_err_2}
    \end{subfigure}
    \hfill
    \begin{subfigure}[t]{0.23\linewidth}
        \centering
        \includegraphics[width=\linewidth]{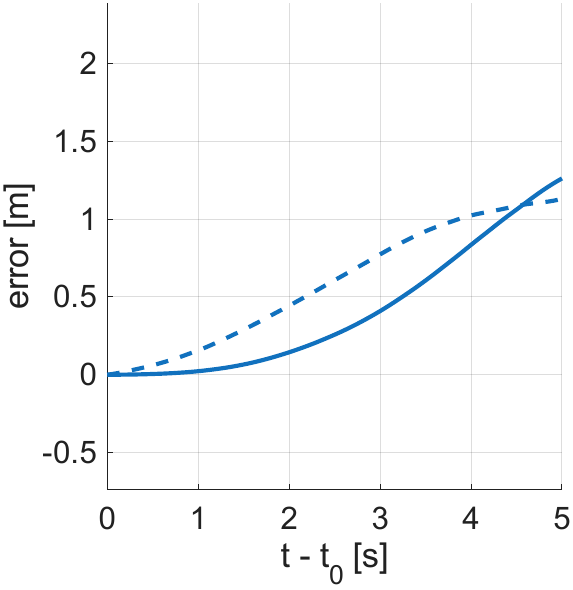}
        \caption{}
        \label{fig:paperclip_err_3}
    \end{subfigure}
    \caption{Top row: measured trajectory (\textcolor[rgb]{0.520,0.520,0.520}{\rule[0.5ex]{11pt}{1.5pt}}) and kinematic model predictions (\textcolor[rgb]{0.1216,0.4667,0.7059}{\hdashrule[0.5ex]{11pt}{1.5pt}{2pt 1pt}}) of S$_6$ for four initial times $t_0 \in \{1, 10, 19, 28\}\,\mathrm{s}$, together with the corresponding measured segment $[t_0,\, t_0 + T_{\text{pred}}]$ (\textcolor[rgb]{0,1.0,0}{\rule[0.5ex]{11pt}{1.5pt}}). Bottom row: longitudinal (\textcolor[rgb]{0.1216,0.4667,0.7059}{\rule[0.5ex]{11pt}{1.5pt}}) and lateral (\textcolor[rgb]{0.1216,0.4667,0.7059}{\hdashrule[0.5ex]{11pt}{1.5pt}{2pt 1pt}}) errors over the prediction horizon for the four initial times of the top row.} 
    \label{fig:paperclip_kinematic}
\end{figure*}

\section{Comparison of Simulation Models Against Measurements} \label{sec:results}
To quantify the short-horizon position accuracy of the vehicle models presented in \Section\ref{sec:veh_mod}, the trajectory measured by the OxTS mounted in the Nissan Leaf serves as ground truth across all scenarios. The following scenarios are considered:
\begin{enumerate}[label=S$_\arabic*$:, leftmargin=*, labelwidth=2.2em, labelsep=0.5em, align=left]
    \item Figure-eight at an approximately constant speed of $4$~m/s.
    \item[S$_2$-S$_4$:] Steady-state cornering at approximately $1.5$, $4$, and $6$~m/s$^2$ lateral acceleration ($R \approx 23$~m).
    \item[S$_5$:] Steady-state cornering at $a_y \approx-1.5$~m/s$^2$ ($R \approx 20$~m), followed by deceleration to standstill.
    \item[S$_6$:] Right U-turn, straight segment, left U-turn, straight segment, and left U-turn, decelerating to approximately $4$~m/s before each turn and accelerating to $6$~m/s afterward.
    \item[S$_7$:] Driving at $16$~m/s with a slight right turn, decelerating to $5$~m/s, driving a $450\degree$ roundabout, and accelerating back to $16$~m/s on exit.
\end{enumerate}

At $100$~ms intervals, the three models are initialized with the measured state and simulated over a prediction horizon $T_{\text{pred}}=5$~s using fourth-order Runge-Kutta with a $50$~ms step. The result is illustrated for the kinematic model in the top row of \Figure\ref{fig:paperclip_kinematic}, showing the full measured trajectory of S$_6$ alongside the predicted and measured segments $[t_0,\,t_0+T_{\text{pred}}]$ for $t_0 \in \{1, 10, 19, 28\}\,\mathrm{s}$. From the simulated trajectory, the predicted OxTS position is obtained from the \CG~position using \eqref{eq:pos_OxTS}.~The predicted and measured OxTS positions are then compared by decomposing the global error $\bm{e}_{\text{glob}}(t) = \bm{p}_{\text{pred}}(t) - \bm{p}_{\text{meas}}(t)$, where $\bm{p}(t) = \left[x_{\text{OxTS}}(t),\,y_{\text{OxTS}}(t)\right]^\top$, into longitudinal and lateral components by rotating by the measured heading $\psi_{\text{meas}}(t)$:
\begin{equation}
    \begin{bmatrix} e_{\text{lon}}(t) \\ e_{\text{lat}}(t) \end{bmatrix} = \begin{bmatrix} \phantom{-}\cos\left(\psi_{\text{meas}}(t)\right) & \sin\left(\psi_{\text{meas}}(t)\right) \\ -\sin\left(\psi_{\text{meas}}(t)\right) & \cos\left(\psi_{\text{meas}}(t)\right) \end{bmatrix} \bm{e}_{\text{glob}}(t),
\end{equation}
where $e_{\text{lon}}$ and $e_{\text{lat}}$ are the longitudinal and lateral errors, respectively. The bottom row of \Figure\ref{fig:paperclip_kinematic} shows the resulting longitudinal and lateral errors of the kinematic model for $t_0 \in \{1, 10, 19, 28\}\,\mathrm{s}$ of S$_6$ over the prediction horizon. Considering the vehicle coordinate frame, where the longitudinal axis points forward and the lateral axis points left, positive longitudinal and lateral errors indicate that the predicted position is ahead and to the left of the measured position, respectively.

\begin{figure}[t]
    \centering
    \includegraphics[width=0.46\columnwidth]{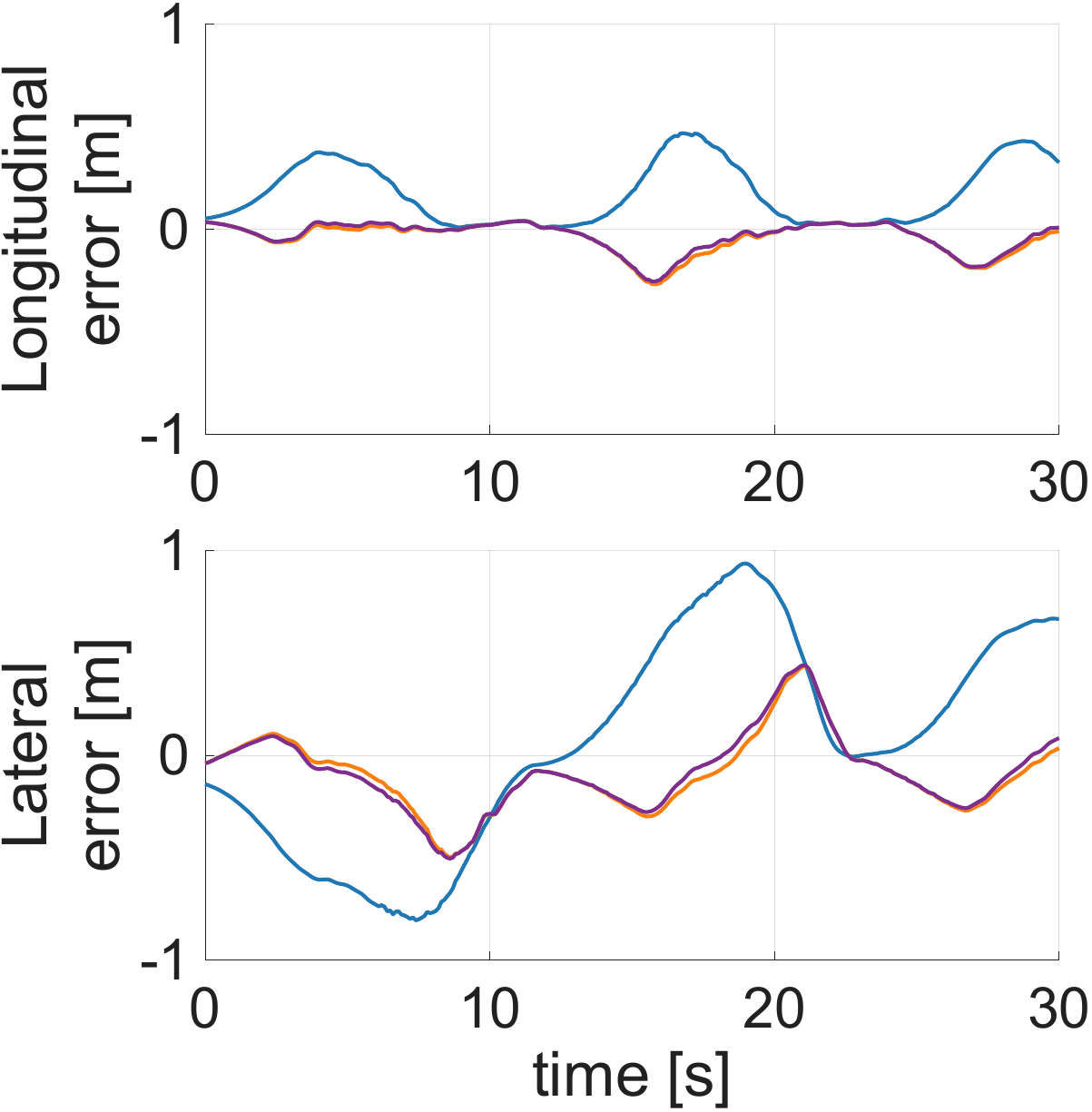}
    \caption{Longitudinal and lateral errors averaged over each prediction horizon $[t_0,\,t_0+T_{\text{pred}}]$ as functions of $t_0$ for the kinematic (\textcolor[rgb]{0.1216,0.4667,0.7059}{\rule[0.5ex]{11pt}{1.5pt}}), linear dynamic (\textcolor[rgb]{1.0000,0.4980,0.0549}{\rule[0.5ex]{11pt}{1.5pt}}), and nonlinear dynamic (\textcolor[rgb]{0.4941,0.1843,0.5569}{\rule[0.5ex]{11pt}{1.5pt}}) models for S$_6$.}
    \label{fig:MPC_error_small}
\end{figure}

{Subsequently, for each initialization time $t_0$ and each of the three models, the longitudinal and lateral errors are averaged over the corresponding prediction horizon $[t_0,\, t_0+T_{\text{pred}}]$. The result is shown in \Figure\ref{fig:MPC_error_small}, where the linear and nonlinear dynamic models show comparable accuracy, whereas the kinematic model shows errors approximately three times as large from the measured position during cornering.}

This approach is applied to each driving scenario. The results are summarized in \Table\ref{tab:err_mean_rmse}, which reports the mean error, maximum absolute error, and RMSE of the longitudinal and lateral trajectory errors over each prediction horizon $[t_0,\,t_0+T_{\text{pred}}]$, for each scenario and model. The mean error indicates systematic bias, the maximum absolute error reflects the worst-case deviation, and the RMSE captures overall magnitude while penalizing larger deviations more heavily. Note that longitudinal and lateral errors are inherently coupled when trajectories are compared in the time domain rather than the spatial domain: if the modeled trajectory follows a higher curvature through a steady-state corner while the velocity profile is accurately tracked, the resulting lateral deviation simultaneously induces a longitudinal error.

\begin{table}[t]
\centering
\caption{Mean error, maximum absolute error and root mean square error (RMSE) of the longitudinal and lateral errors over each prediction horizon $[t_0,\,t_0+T_{\text{pred}}]$ per scenario and model, with m$_1$, m$_2$ and m$_3$ denoting the kinematic, linear dynamic and nonlinear dynamic model, respectively. Bold values indicate the lowest error magnitude per scenario, error type, and direction (longitudinal/lateral).}
\label{tab:err_mean_rmse}
\begin{tabular}{cccc}
\multirow{2}{*}{\textbf{}} & \multirow{2}{*}{\textbf{}} &
\multicolumn{2}{c}{\textbf{Error (mean $\mid$ max $\mid$ RMSE) [m]}} \\
\cmidrule(lr){3-4}
& & \textbf{Longitudinal} & \textbf{Lateral} \\
\midrule
\multirow[c]{3}{*}{S$_1$}
& m$_1$ & $\phantom{-}0.10 \mid 0.19 \mid 0.12$                            & $-0.09 \mid 0.49 \mid 0.33$ \\
& m$_2$ & $\phantom{-}0.02 \mid \mathbf{0.07} \mid \mathbf{0.04}$           & $\mathbf{-0.06} \mid \mathbf{0.26} \mid \mathbf{0.13}$ \\
& m$_3$ & $\phantom{-}\mathbf{0.01} \mid \mathbf{0.07} \mid \mathbf{0.04}$  & $\mathbf{-0.06} \mid \mathbf{0.26} \mid \mathbf{0.13}$ \\
\midrule
\multirow[c]{3}{*}{S$_2$}
& m$_1$ & $\phantom{-}0.17 \mid 0.32 \mid 0.18$                            & $\phantom{-}0.38 \mid 0.63 \mid 0.41$ \\
& m$_2$ & $\phantom{-}\mathbf{0.00} \mid \mathbf{0.09} \mid \mathbf{0.02}$  & $\phantom{-}0.01 \mid \mathbf{0.34} \mid \mathbf{0.06}$ \\
& m$_3$ & $\phantom{-}\mathbf{0.00} \mid \mathbf{0.09} \mid \mathbf{0.02}$  & $\phantom{-}\mathbf{0.00} \mid \mathbf{0.34} \mid \mathbf{0.06}$ \\
\midrule
\multirow[c]{3}{*}{S$_3$}
& m$_1$ & $\phantom{-}0.85 \mid 1.02 \mid 0.87$                            & $\phantom{-}1.20 \mid 1.41 \mid 1.21$ \\
& m$_2$ & $\mathbf{-0.21} \mid \mathbf{0.30} \mid \mathbf{0.22}$            & $\mathbf{-0.30} \mid \mathbf{0.37} \mid \mathbf{0.30}$ \\
& m$_3$ & $-0.25 \mid 0.35 \mid 0.25$                                       & $-0.33 \mid 0.41 \mid 0.34$ \\
\midrule
\multirow[c]{3}{*}{S$_4$}
& m$_1$ & $\phantom{-}2.49 \mid 3.05 \mid 2.56$                            & $\phantom{-}2.88 \mid 3.41 \mid 2.94$ \\
& m$_2$ & $\mathbf{-0.45} \mid \mathbf{0.61} \mid \mathbf{0.46}$            & $\mathbf{-0.46} \mid \mathbf{0.59} \mid \mathbf{0.46}$ \\
& m$_3$ & $-0.77 \mid 0.98 \mid 0.79$                                       & $-0.67 \mid 0.77 \mid 0.68$ \\
\midrule
\multirow[c]{3}{*}{S$_5$}
& m$_1$ & $\phantom{-}0.17 \mid 0.47 \mid 0.23$                            & $\phantom{-}\mathbf{0.05} \mid 0.94 \mid 0.52$ \\
& m$_2$ & $\mathbf{-0.04} \mid \mathbf{0.27} \mid \mathbf{0.09}$            & $-0.09 \mid 0.50 \mid \mathbf{0.20}$ \\
& m$_3$ & $-0.05 \mid 0.28 \mid \mathbf{0.09}$                              & $-0.09 \mid \mathbf{0.49} \mid \mathbf{0.20}$ \\
\midrule
\multirow[c]{3}{*}{S$_6$}
& m$_1$ & $\phantom{-}0.21 \mid 0.26 \mid 0.22$                            & $-0.48 \mid 0.54 \mid 0.48$ \\
& m$_2$ & $\phantom{-}\mathbf{0.07} \mid \mathbf{0.10} \mid \mathbf{0.07}$  & $\mathbf{-0.16} \mid \mathbf{0.19} \mid \mathbf{0.16}$ \\
& m$_3$ & $\phantom{-}\mathbf{0.07} \mid \mathbf{0.10} \mid \mathbf{0.07}$  & $\mathbf{-0.16} \mid \mathbf{0.19} \mid \mathbf{0.16}$ \\
\midrule
\multirow[c]{3}{*}{S$_7$}
& m$_1$ & $\phantom{-}0.09 \mid 0.63 \mid 0.20$                            & $\mathbf{-0.03} \mid 2.63 \mid 0.83$ \\
& m$_2$ & $\mathbf{-0.02} \mid \mathbf{0.11} \mid \mathbf{0.04}$            & $\phantom{-}\mathbf{0.03} \mid \mathbf{0.59} \mid \mathbf{0.14}$ \\
& m$_3$ & $\mathbf{-0.02} \mid 0.12 \mid \mathbf{0.04}$                     & $\phantom{-}\mathbf{0.03} \mid 0.60 \mid 0.15$ \\
\bottomrule
\end{tabular}
\end{table}

The two dynamic models generally outperform the kinematic model by a factor of two to six for both longitudinal and lateral errors in most scenarios, as expected given their ability to capture tire side slip and vehicle body dynamics. At higher lateral accelerations, the dynamic behavior of the vehicle has a more significant influence on the trajectory, as the kinematic model neglects tire slip and therefore cannot reproduce speed-dependent effects such as understeer. This is most pronounced in the steady-state cornering scenarios S$_3$ ($a_y \approx 4~\text{m/s}^2$) and S$_4$ ($a_y \approx 6~\text{m/s}^2$). In both scenarios, the vehicle drives a counterclockwise circular path, i.e., a left circle. The positive mean lateral error of the kinematic model indicates that the predicted trajectory has a smaller radius than the reference, confirming that understeer is not captured by the kinematic model.

The linear and nonlinear dynamic models show comparable accuracy for all scenarios, with relative differences in the centimeter range (except S$_4$) and absolute errors with respect to the measured trajectories in the decimeter range. The similarity stems from two factors. Firstly, the tire slip angles remain largely within or close to the linear region. For all scenarios except S$_4$, the front slip angle $\alpha_0$ is within $\pm 3^\circ$, the rear slip angle $\alpha_1$ stays between $\pm 2^\circ$. Since the linear and nonlinear force responses are aligned via $\lambda_{K y \alpha i}$ as visualized in \Figure\ref{fig:comparison_linear_nonlinear_model}, both tire-force formulations yield near-identical lateral force responses under these conditions. Secondly, the longitudinal acceleration in the tested scenarios is limited, resulting in little load transfer effects. This also justifies the use of pure side slip rather than combined slip in the current Magic Formula implementation. 

\begin{figure}
    \centering
    \includegraphics[width=0.48\columnwidth]{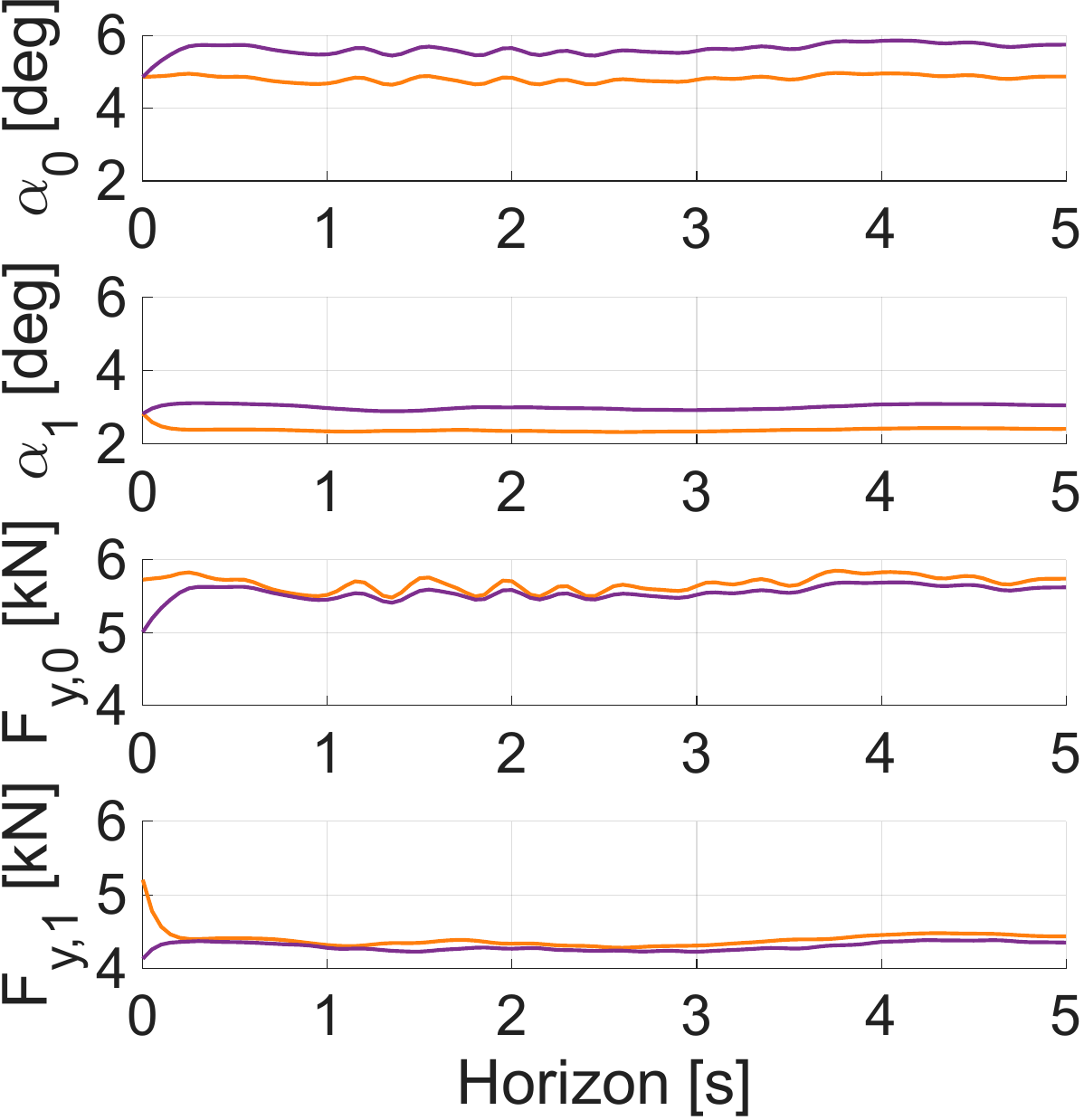}
    \caption{Predicted side slip angles and corresponding lateral forces for the front and rear tires for the linear dynamic (\textcolor[rgb]{1.0000,0.4980,0.0549}{\rule[0.5ex]{11pt}{1.5pt}}) and nonlinear dynamic (\textcolor[rgb]{0.4941,0.1843,0.5569}{\rule[0.5ex]{11pt}{1.5pt}}) models for S$_4$ at $t_0 = 25$\,s.}
    \label{fig:roundabout_3_slipangles}
\end{figure}

For scenario $\text{S}_4$, the predicted side slip angles of the linear and nonlinear dynamic models differ over the prediction horizon, as shown in \Figure\ref{fig:roundabout_3_slipangles}, where both models are initialized from identical states. However, at the higher slip angles that occur in this scenario, the Magic Formula yields lower lateral forces than the linear approximation due to saturation at higher slip levels. This causes the slip angles of the linear and nonlinear model to settle at different steady-state values, with a lateral force difference of approximately 100~N at both axles. The linear model achieves lower trajectory error than the nonlinear model, contrary to the expectation that increased model fidelity would improve the position accuracy. Both models give a negative average error for the counterclockwise circle of S$_4$, which means that the predicted trajectories are to the outside of the circle. The reason for this is twofold: the identified tire parameters and neglecting steering compliance and steering dynamics.

As described in \Section\ref{sec:vehicle_measurement_platform}, the cornering stiffness $C_{\alpha i}$ is identified from experimental data via a least-squares fit of the linear tire model for $|a_y| \leq 4$~m/s$^2$. The cornering stiffness of the nonlinear tire model is then scaled to match the identified $C_{\alpha i}$, which also affects the lateral force response at higher slip angles. However, at these higher slip levels, the available lateral force is primarily determined by the available friction. Since the friction level during MF-parameter identification is assumed equal to and constant across all driving scenarios, the modeled nonlinear force response may deviate from the true tire behavior. This highlights the importance of accurately characterizing operating conditions, and underscores a fundamental challenge of higher-fidelity models. Their typically larger number of parameters multiplies the potential sources of identification uncertainty, each of which can degrade their accuracy.

At the end of S$_5$, the vehicle brakes to a standstill at $\vxin \approx -1.5$~m/s$^2$, producing approximately 400~N of longitudinal load transfer according to \eqref{eq:load_transfer}. The resulting front and rear slip angles are $\alpha_0 \approx 1^\circ$ and $\alpha_1 \approx 0.5^\circ$, at which the change in lateral tire force due to load transfer is approximately 50~N. Consequently, the center-of-gravity height has negligible influence in this scenario given the low longitudinal accelerations, and the linear and nonlinear dynamic models predict comparable trajectories.

\section{Discussion} \label{sec:discussion}
The three single-track model variants evaluated in \Section \ref{sec:results} represent a common starting point for MPC design in autonomous driving. Depending on the operating conditions and vehicle type, several additional modeling effects may be considered. The following list highlights the most relevant ones and discusses when and why they become important.

\begin{itemize}
    \item \textbf{Steering dynamics:} The wheel steering angle is derived from the steering wheel angle via a constant ratio using~\eqref{eq:steering}, neglecting steering system compliance and tire force feedback. Thus, at higher lateral accelerations and when the MPC actuates the steering wheel directly, modeling the steering dynamics may be necessary.
    \item \textbf{Longitudinal load transfer:} The effect is limited at moderate longitudinal accelerations but becomes more significant at higher ones. It also varies with loading conditions: for vehicles with variable payload such as taxis or trucks, the \CG~position shifts accordingly.
    \item \textbf{Combined slip:} Simultaneous longitudinal and lateral force generation reduces the available force, as bounded by the friction ellipse. This is relevant during braking in a turn and can be addressed through a combined slip formulation or explicit friction ellipse constraints.
    \item \textbf{Aerodynamic forces:} At high speeds, drag and downforce affect longitudinal dynamics and vertical tire loads. Crosswind introduces lateral forces, which is particularly relevant for large, high-sided vehicles such as trucks and buses.
    \item \textbf{Low-velocity singularity:} The slip angle formulation in~\eqref{eq:sideslipangles} requires a regularization term $v_\epsilon$ to avoid singularities near standstill. A relaxation-length tire model avoids this at the cost of one additional state per axle~\cite{2012_Pacejka_Tirevehicledynamics}.
    \item \textbf{Lateral load transfer:} In cornering, lateral acceleration shifts vertical load to the outer wheels. The single-track model does not capture this but a double-track model can represent it at the cost of additional states and parameters.
\end{itemize}

In addition to the possible modeling variants, a practical constraint in MPC is that the optimal control problem must be solved within a given sampling interval. Increasing model fidelity raises the computational cost of both evaluating the prediction model and solving the resulting optimization problem, which can limit the achievable prediction horizon or require a longer sampling interval. In closed-loop operation, this trade-off is consequential: a shorter horizon reduces the lookahead distance available to the controller, while a longer sampling interval reduces responsiveness of the autonomous vehicle to disturbances. Therefore, the effect of model fidelity on the closed-loop performance is worth considering, such as investigated by Subosits and Gerdes~\cite{10.1109/TIV.2021.3051325} for maneuvers at friction limits.

The results together with the discussion indicate that the predictive accuracy of any model is bounded by the model complexity in combination with the quality of its parameter identification. Poorly identified parameters can offset the theoretical advantage of increasing model fidelity, as substantiated by the results of the linear and nonlinear models in this work. This aligns with Box's aphorism that \textit{all models are wrong, but some are useful} \cite{10.1080/01621459.1976.10480949}: selecting the appropriate level of fidelity based on the expected operating conditions and ensuring that the corresponding parameters are accurately identified are therefore at least as important as the choice of model structure itself.

\section{Conclusion} \label{sec:conc}

This paper evaluates the short-horizon position accuracy of three single-track vehicle models (kinematic, linear dynamic, and nonlinear dynamic) against measurements from a Nissan Leaf ZE0 across a range of driving maneuvers, together with the aim of bridging the gap between the vehicle dynamics and motion planning communities by guiding the choice of model and its parameterization for MPC applications in autonomous driving.

The kinematic model performs substantially worse than both dynamic models, by approximately a factor of two to six in most scenarios, as it neglects tire slip and therefore cannot reproduce speed-dependent effects such as understeer. This is most pronounced in the steady-state cornering scenarios at $a_y \approx 4$~m/s$^2$~and $a_y \approx 6$~m/s$^2$. The linear and nonlinear dynamic models achieve comparable accuracy across all scenarios, since tire slip angles remain largely within the linear regime and longitudinal load transfer effects are small for the test cases considered. In the steady-state cornering scenario at $a_y \approx 6$~m/s$^2$, the nonlinear model unexpectedly yields larger errors than the linear model, indicating that model accuracy is bounded by the quality of parameter identification. This highlights that increasing model complexity does not inherently improve position accuracy if the additional parameters cannot be accurately identified.

Therefore, the linear dynamic model offers the best trade-off between accuracy and parameterization effort for the scenarios considered. The kinematic model is straightforward to parameterize but lacks accuracy at higher lateral accelerations. The nonlinear dynamic model requires greater parameterization effort yet does not yield higher accuracy, as the test scenarios remain largely within the linear tire regime and longitudinal load transfer effects are limited.

Future work entails implementing these models in an MPC framework on an autonomous vehicle to assess which model provides sufficient accuracy in closed-loop applications and to evaluate the associated computational cost. The dataset should be extended with more diverse driving maneuvers at higher speeds and lateral accelerations to better characterize the transition to nonlinear tire behavior, including combined slip conditions and longitudinal load transfer. Under such conditions, the effects of steering dynamics, tire relaxation, lateral load transfer, and aerodynamic forces are expected to become more significant and should be investigated. Finally, this work should be extended to autonomous trucks, where payload variability, a higher center of gravity, and structural flexibility such as trailer chassis bending introduce additional modeling challenges. Furthermore, the required prediction horizon for trucks is considerably longer than for passenger cars, as larger vehicle dimensions and limited braking performance necessitate greater lookahead distances. Since model errors accumulate over the prediction horizon, positional accuracy is expected to degrade more significantly at the longer horizons required for autonomous trucks.

\section{Acknowledgments}
This work has been carried out within the EU-funded Horizon Europe project MODI (grant agreement ID: 101076810). 

All authors reviewed the manuscript draft and revised it critically on intellectual content. All authors approved the final version of the manuscript to be published.

Conflicts of Interest: None

\bibliographystyle{IJAEtran}
\bibliography{2026_AVEC26}

\appendix
\section{Lateral Force Magic Formula equations} \label{sec:appendix_MF}
This appendix provides the lateral force Magic Formula equations for pure side slip situations with vertical force dependency, i.e., without pressure, camber dependencies or offsets. The Magic Formula equations for pure slip lateral force are given by \cite{2012_Pacejka_Tirevehicledynamics}:
{\allowdisplaybreaks
\begin{subequations} \label{eq:MF}
\begin{align}
    F_{y} = &D_y \sin\!\left(C_y \arctan\left(B_y\alpha_y\right.\right. \nonumber\\ 
    &\left.\left. - E_y\left(B_y\alpha_y - \arctan\left(B_y\alpha_y\right)\right)\right)\right) + S_{Vy}, \\
    C_y &= p_{Cy1}\lambda_{Cy}\\
    D_y &= \mu_y F_z, \\
    \mu_y &= \left(p_{Dy1} + p_{Dy2}\,df_z\right)\,\lambda_{\mu y}, \\
    K_{y\alpha} &= p_{Ky1}F_{z0}\sin\!\left(p_{Ky4}\arctan\!\left(\frac{F_z}{p_{Ky2}F_{z0}}\right)\right)\lambda_{Ky\alpha}, \label{eq:MF_cornering_stiffness} \\
    B_y &= \frac{K_{y\alpha}}{C_y D_y},\\
    E_y &= \left(p_{Ey1} + p_{Ey2}\,df_z\right)\left(1 - p_{Ey3}\,\text{sgn}\left(\alpha_y\right)\right)\\
    df_z &= \frac{F_z - F_{z0}}{F_{z0}}.
\end{align}
\end{subequations}}

The Magic Formula parameters for the front and rear tires of the Nissan Leaf are given in \Table \ref{tab:MF_pars}.

\begin{table}[h]
    \centering
    \caption{Lateral force Magic Formula parameters of the front tire (Continental Conti Premium Contact 5 205/55 R16 91V) and rear tire (Michelin Primacy 4 205/55R16 91H). Note that the tire is assumed to be symmetric in the simulations such that $F_{yi} = 0$~N for $\alpha_i = 0^\circ$. In other words, $p_{hy1}=p_{hy2}=p_{vy1}=p_{vy2}=0$. The Magic Formula parameters follow the SAE sign convention, in which $p_{Ky1}<0$ yields a negative cornering stiffness $K_{y\alpha i}$.}
    \label{tab:MF_pars}
    \begin{tabular}{c|c|c}
         \textbf{Parameter} & \textbf{Front tire $(i=0)$} & \textbf{Rear tire $(i=1)$} \\
         \hline
         $F_{z0}$ & 4250~N & 3300~N \\
         $p_{Cy1}$ & 1.05    & 1.20    \\
         $p_{Dy1}$ & 1.02    & 0.986   \\
         $p_{Dy2}$ & -0.148  & -0.112  \\
         $p_{Ey1}$ & -0.463  & 0.421   \\
         $p_{Ey2}$ & -2.00   & -2.00   \\
         $p_{Ey3}$ & 0.00   & 0.00   \\
         $p_{Ky1}$ & -20.0   & -23.7   \\
         $p_{Ky2}$ & 1.05    & 2.50    \\
         $p_{Ky4}$ & 1.72    & 2.00    \\
         $\lambda_{Ky\alpha i}$ & 0.423 &  0.959
    \end{tabular}
\end{table}

\end{document}